%% file: MDRpaper.tex
\documentclass{article}

\usepackage{times}
\usepackage{graphicx} %
\usepackage{subfigure} 

\usepackage{booktabs}

\usepackage{natbib}

\usepackage{algorithm}
\usepackage{algorithmic}

\usepackage{hyperref}

\usepackage[accepted]{icml2016}

\icmltitlerunning{Maximally Divergent Intervals for Anomaly Detection}

\usepackage{amssymb,amsmath,amstext,amsbsy,multirow,color,amsthm}
 \usepackage{amsthm}
\usepackage{bm}
\usepackage{xspace}
\DeclareMathOperator*{\argmax}{argmax}

\newcommand\eg{\emph{e.g.}\xspace}
\newcommand\ie{\emph{i.e.}\xspace}

\begin{document} 

\twocolumn[
\icmltitle{Maximally Divergent Intervals for Anomaly Detection}

\icmlauthor{Erik Rodner$^{1,3}$, Bj\"orn Barz$^1$, Yanira Guanche$^1$,  Milan Flach$^2$, Miguel Mahecha$^{2,3}$, Paul Bodesheim$^2$, Markus Reichstein$^{2,3}$, Joachim Denzler$^{1,3}$} {erik.rodner@uni-jena.de}
\icmladdress{\vspace{3pt}$^1$Computer Vision Group,  Friedrich Schiller University of Jena, Germany, \url{http://www.inf-cv.uni-jena.de}}
\icmladdress{\vspace{-8pt}$^2$ Max Planck Institute for Biogeochemistry, Department Biogeochemical Integration, Jena, Germany}
\icmladdress{\vspace{-8pt}$^3$ Michael Stifel Center for Data-driven and Simulation Science, Jena, Germany}

\icmlkeywords{Kullback-Leibler divergence, anomaly detection, extreme event detection, novelty detection}

\vskip 0.3in
]

\begin{abstract} 
    We present new methods for batch anomaly detection in multivariate time series.
    Our methods are based on maximizing the Kullback-Leibler divergence between the data distribution within and outside an interval of the time series. 
    An empirical analysis shows the benefits of our algorithms compared to methods
    that treat each time step independently from each other without optimizing with respect to all possible intervals.
\end{abstract} 

\vspace{-3mm}
\section{Introduction}\label{intro}

Scientific data increases with respect to both volume and dimensionality. 
Manually analyzing large-scale multivariate data is therefore intractable and automatic methods are needed to structure data and point researchers to the most interesting parts of scientific measurements. We focus on detecting anomalies in time series, which is an essential task, \eg, in climate and ecosystem studies \cite{zscheischler2014GPPextr}, oceanic research \cite{minguez2012regression}, or in industrial processes \cite{darkow2014indextr}.

The survey article of \cite{chandola2009anomaly} categorizes anomaly event detection methods into six main groups. \textit{Classification-based} methods utilize common classification techniques, such as neural networks, Bayesian networks \cite{dieh2002real}, or Support Vector Machines \cite{ma2003time} and learn their models by sliding windows strategies. Anomalies can also be detected by considering the distance to the $k^{th}$ \textit{nearest neighbor} \cite{byers1998nearest,knfst} 
or the relative density \cite{chiu2003enhancements}. \textit{Clustering} techniques~\cite{smith2002clustering} group similar data into clusters leading to anomalies being far from the cluster centroids \cite{smith2002clustering}. 
An intuitive strategy is based on \textit{statistical modeling}, 
where anomalies are assumed to be points that do not fit to a previously estimated statistical model \cite{anscombe1960rejection,minguez2012regression}. 
Other approaches are based on \textit{information theory}~\cite{ando2007clustering,bmvc} or 
\textit{spectral analysis}~\cite{shyu2003novel}, where subspaces of the normal data are detected.

All of these methods determine an anomaly score for each point in the time series individually. In contrast, we propose a method that directly considers
the detection of contiguous intervals, an important property for scientific data analysis. Time intervals whose distribution is considerably different from the rest of the time series can be considered as anomalies. This is done by maximizing a divergence criterion between the distributions.
Depending on the assumptions on these distributions, we derive different methods that allow for batch detection of anomalies.

The most related paper to ours is the method proposed by \cite{liu2013change}, which also uses a divergence criterion to detect changes in the data. However, their method makes use of the more general $f$-divergence and directly estimates the ratio
of the two densities. Since we are optimizing over all possible intervals instead of only a window of fixed size, we rely on efficient update formulas, which are not available for the methods proposed in \cite{liu2013change}.
The paper of \cite{goernitz2015hidden} combines a hidden Markov model with a latent one-class SVM for detecting time series containing an anomaly. Their method requires some kind of supervision to learn a state model and also does not directly focus on anomaly localization
in contrast to our approach.

In the following sections, we give a brief description of the problem, the general framework we propose 
for batch anomaly detection as well as specific algorithms.
Subsequent experiments show the properties and benefits of the algorithms especially with respect to single point
anomaly detection. Furthermore, we propose average precision and an intersection-over-union criterion as
a suitable evaluation methodology for anomaly detection in time series.

\vspace{-3mm}
\section{Maximally divergent intervals (MDI)} \label{section:2}

\paragraph{Definitions and problem description}
The basic idea behind our approach is that an anomaly interval within a data distribution is significantly different from the rest of the time series. Therefore our main objective is to be able to find intervals in a time series $(\bm{x}_t)_{t=1}^n$ with $\bm{x}_t \in \mathbb{R}^D$ being a multivariate observation at time $t$. 

Let $I = \{t \;|\; t_1 \leq t < t_2\}$ be an interval, where data points are assumed to be sampled from $p_I$. The remaining set of data points is denoted by $\Omega = \{1, \ldots, n\} \setminus I$ with the data distributed by $p_{\Omega}$. In order to find those intervals: i) we need to define a parameterized model for the distributions $p_I$
and $p_{\Omega}$ that can be estimated from the data, and ii) be able to calculate the ``difference'' between $p_I$ and $p_{\Omega}$. 

For the latter, we propose to use the Kullback-Leibler (KL) divergence to measure the difference between distributions. Furthermore, we model the data distributions either by kernel density estimation (KDE) or multivariate Gaussian distributions. These two models allow us to compute the KL divergence in an efficient manner. 

\vspace{-2.5mm}
\paragraph{Maximizing the Kullback-Leibler divergence}
In the following, we assume the data points in either $\Omega$ and $I$ to be sampled independently from each other.
This is for sure a severe assumption that does not hold for relevant time series, however, we will demonstrate
later on how the dependencies between subsequent data points can be handled with a simple pre-processing step.
The Kullback-Leibler divergence of two distributions $p_{\Omega}$ and $p_I$ is defined as: 
\begin{align}
    \text{KL}(p_I, p_{\Omega}) &= \int p_I(\bm{x}) \log \frac{ p_{I}(\bm{x}) }{ p_{\Omega}(\bm{x}) } \mathrm{d}\bm{x}\enspace.
\end{align}
The KL divergence is zero for identical distributions and large for ``significantly different'' data distributions.
We approximate it using an empirical expectation over the set of anomalous points leading to:
\begin{align}
    \text{KL}_{I, \Omega} 
    &= \frac{1}{|I|} \sum\limits_{t \in I} \left( \log p_{I}(\bm{x}_t) - \log p_{\Omega}(\bm{x}_t)\right)
\end{align}
This resulting criterion is very intuitive since it is calculating the differences of log-likelihoods for $p_{I}$ and $p_{\Omega}$.
To find the interval belonging to an anomaly, we maximize the KL divergence with respect to the interval $I$:
\begin{align}
    \hat{I} &= \textstyle \argmax_{I \in \mathcal{I}}\; \text{KL}_{I, \Omega} \enspace.
\end{align}
The set $\mathcal{I}$ contains suitable intervals and is important to integrate prior expectations about anomaly intervals, such as a range of possible interval sizes. Naive brute-force optimization of the empirical KL divergence 
requires $\mathcal{O}(|\mathcal{I}| \cdot T)$ operations, where $T$ is the time needed to evaluate the KL divergence 
and $\mathcal{I}$ is usually $\mathcal{O}(n \cdot n')$ with $n'$ being the maximum size of an anomaly interval.
A property of the KL divergence is its asymmetry,  $\text{KL}_{I, \Omega} \neq \text{KL}_{\Omega, I}$. 
Other work~\cite{liu2013change} often relied on a symmetric version of it. We use $\text{KL}_{I, \Omega}$ for
reasons theoretically explained later on and validated in our experiments.
To obtain $m$ anomalies, a non-maximum-suppression method \cite{neubeck2006efficient} is used to select the $m$ non-overlapping intervals with highest divergence.

\vspace{-2.5mm}
\paragraph{MDI with kernel density estimation (MDI KDE)}
A very flexible way to model and estimate distributions is kernel density estimation (KDE).
For a given kernel function $K$, the estimate for $p_I$ is defined by:
\begin{align}
p_I(\mathbf{x}) &= \textstyle \frac{1}{|I|} \sum_{t_1 \leq t < t_2 } K(\mathbf{x}, \mathbf{x}_t)
\end{align}
for an arbitrary multivariate observation $\mathbf{x}$. We use the same model for $p_{\Omega}$. 
As a kernel function, we use the Gaussian kernel normalized such that $p_I$ is a proper density.

Straightforward computation of the KL divergence for $p_I$ and $p_{\Omega}$ estimated by kernel density estimation
requires $O(n^2)$ operations (distance calculations). Together with our brute-force optimization, 
this yields an $\mathcal{O}(n^3 \cdot n')$ algorithm, which is only practical for small time series.
However, the idea of cumulative sums is used to achieve a significant speed-up \cite{viola2004robust}.
Let $\bm{K} = (K_{t, t'}) \in \mathbb{R}^{n \times n}$ be the kernel matrix of the time series.
We compute the cumulative sums in this symmetric matrix along a single axis:
    $C_{t, t'} = \sum_{t'' \leq t'} K_{t, t''}$
which can be pre-computed in $\mathcal{O}(n^2)$ time. Since we only need to compute our kernel density estimate
for points of the time series, we can evaluate the estimates in constant time:
\begin{align}
    p_I(\bm{x}_t) &= \frac{1}{|I|} \left( C_{t, t_2-1} - C_{t, t_1-1} \right) \enspace,\\
    p_{\Omega}(\bm{x}_t) &= \frac{1}{n - |I|} \left( C_{t, n} - C_{t, t_2-1} + C_{t, t_1-1} \right) \enspace.
\end{align}
After computing the kernel matrix and cumulative sums in $\mathcal{O}(n^2)$ time, we get an asymptotic time of $\mathcal{O}(n')$ for evaluating the KL divergence and a total time of $\mathcal{O}(\max(n^2, n'^2 \cdot n))$ for finding the maximally divergent interval.

\vspace{-2.5mm}
\paragraph{MDI for normally-distributed data (MDI Gaussian)}

Another possibility to model the data distributions is a Gaussian model for $p_I$ and $p_{\Omega}$:
\begin{align}
p_I(\bm{x}) &= \mathcal{N}(\bm{x} \;|\; \bm{\mu}_I, \bm{S}_I), \;
p_{\Omega}(\bm{x}) = \mathcal{N}(\bm{x} \;|\; \bm{\mu}_{\Omega}, \bm{S}_{\Omega})
\end{align}
Estimating the mean vectors and covariance matrices can be also achieved with integral series. 
The exact KL divergence even has a closed form solution~\cite{duchi2007derivations}:
\begin{align}
\notag
\text{KL}_{I, \Omega} &= \frac{1}{2}\bigl( \text{trace}(\bm{S}_{\Omega}^{-1} \bm{S}_{I}) \\
\notag
    &+ (\bm{\mu}_I - \bm{\mu}_{\Omega})^T \bm{S}_{\Omega}^{-1} (\bm{\mu}_{I} - \bm{\mu}_{\Omega})\\
\label{eq:klgaussian}
    &- D + \log(|\bm{S}_{\Omega}|) - \log(|\bm{S}_{I}|) \bigr) \enspace,
\end{align}
which we can use for an evaluation of the divergence in a time independent of $n$, yielding a total
computation time of $\mathcal{O}(n' \cdot n)$.

Depending on further assumptions on the distributions, interesting connections to related techniques can be derived.
For example, when we assume a global shared covariance matrix ($\bm{S} = \bm{S}_I = \bm{S}_{\Omega}$), 
eq.~\eqref{eq:klgaussian} reduces to $\text{KL}_{I, \Omega} = (\bm{\mu}_I - \bm{\mu}_{\Omega})^T \bm{S}^{-1} (\bm{\mu}_{I} - \bm{\mu}_{\Omega})$
resembling a Mahalanobis distance also used in Hotelling's $T^2$ test~\cite{macgregor1995statistical}. Furthermore, we can assume identity matrices
for the covariances, which reduces eq.~\eqref{eq:klgaussian} to the squared Euclidean distance between the means.
The expression in eq.~\eqref{eq:klgaussian} also justifies our choice of $\text{KL}_{I, \Omega}$ instead of $\text{KL}_{\Omega, I}$ 
or a symmetric version~\cite{liu2013change}. For a univariate time series, $\text{KL}_{\Omega, I}$ is given by:
\begin{align}
\notag
&\frac{1}{2}\left(\frac{S_{\Omega}}{S_{I}} + \frac{(\mu_{I} - \mu_{\Omega})^2}{S_{I}} - 1 + \log(S_I) - \log(S_{\Omega})\right),
\end{align}
and we can see that for small values of $S_I$, we get high values of the KL divergence. 
Therefore, a maximization of $KL_{\Omega, I}$ would prefer intervals of low variance. 
In contrast, $KL_{I, \Omega}$ is not affected by this phenomenon since $S_{\Omega}$ is estimated from a large 
portion of the time series.  

\vspace{-2.5mm}
\paragraph{Temporal context with time-delay embedding}
A major drawback of our algorithm so far is the assumption of independent data points in the time series, which is only valid
for trivial academic cases. However, modeling the dependency can be done with a simple transformation of the time series. 
In particular, we use a multivariate time-delay embedding~\cite{smets2009discovering,kantz2004nonlinear}, where the data points of the new time series are the concatenation 
of the last $k$ time steps of the original time series, \ie $\bm{x}_t' = (\bm{x}_t, \bm{x}_{t-1}, \ldots, \bm{x}_{t-k+1})$.
Whereas this embedding leads to a smoothing of the distance matrix for our MDI KDE approach, it allows our MDI Gaussian approach for calculating 
and exploiting correlations between subsequent data points. For example, a change of frequency in the time series, can only be detected
by our methods with a proper embedding, such as time-delay.

Multiple methods for estimating an ``optimal'' value for $k$ have been proposed in the literature for univariate time series \cite{hegger1999practical}. 
We developed a method that allows for optimizing $k$ even for multivariate time series.
However, the preliminary results we obtained when combining this hyperparameter tuning method
with our MDI approach are beyond the scope of this paper.
In our current experiments, we therefore fix $k=3$ for all methods.

\vspace{-3mm}
\section{Experiments} \label{section:exper}

\paragraph{Evaluation criteria, baselines, and implementation details}
Previous methods in the area of anomaly detection, typically return a novelty score for each of the
data points in the time series. Therefore, a common choice for evaluation have been ROC curves and the area
under these curves (AUC). However, our method returns scored time intervals directly and therefore
differs from previous algorithms. We argue that in this case also a proper detection evaluation criterion needs
to be used, which also better reflects the expectations of researchers about an algorithm's performance.
\begin{figure}
    \centering
    \includegraphics[width=0.9\linewidth]{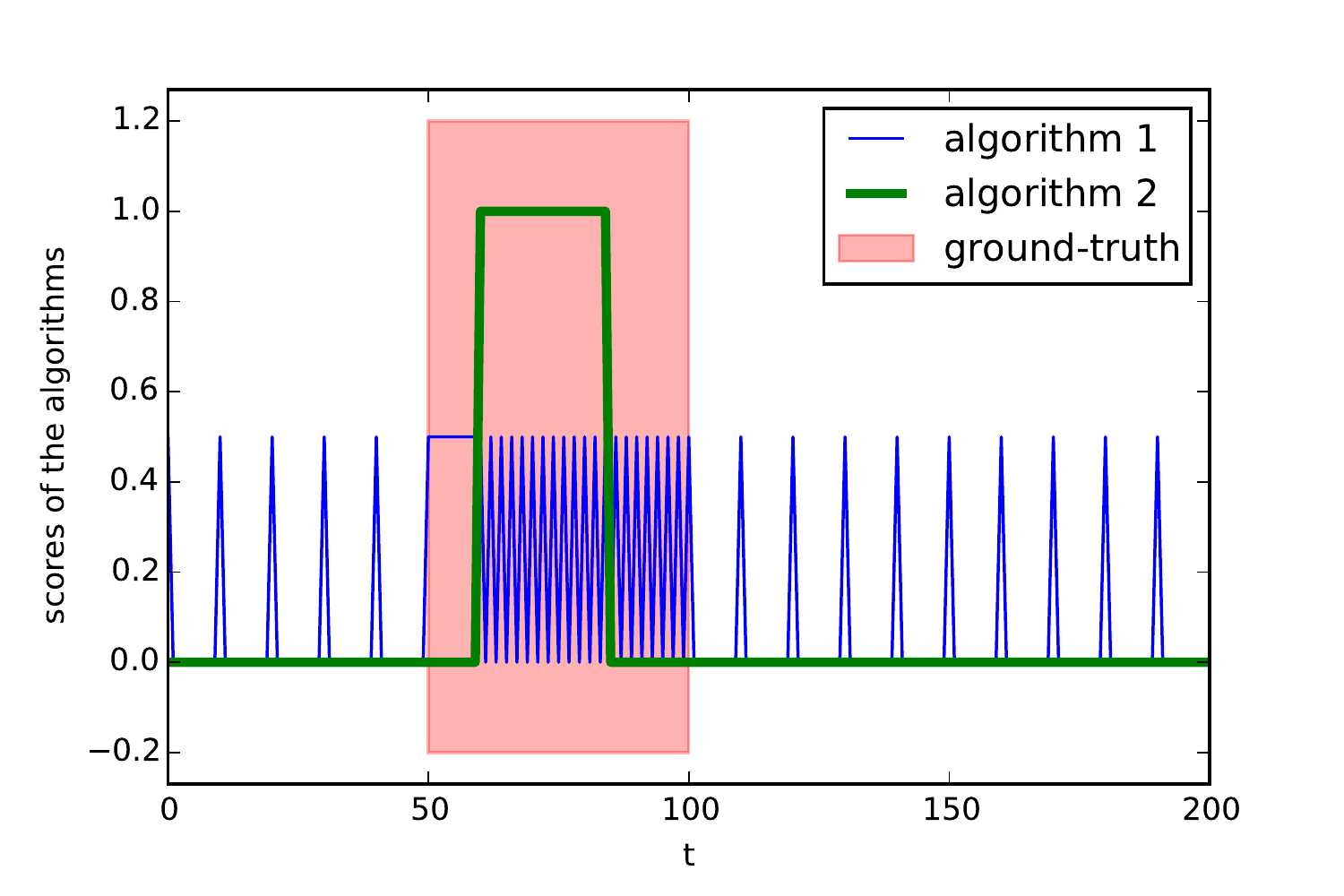}
    \caption{Resulting scores of two algorithms (blue and green) achieving the same AUC performance (0.75) with respect
        to the red ground-truth interval but a significantly different AP detection performance (0.0 for blue vs. 1.0 for green).}
    \label{fig:rocexample}
\end{figure}
Therefore, we count detected intervals as correct if they have an intersection over union ratio of more than $\beta=0.5$ with
a ground-truth anomaly interval. Evaluation is then done using recall-precision curve and the average precision (AP) metric.
Figure~\ref{fig:rocexample} shows an example, where two algorithms have the same AUC but a significantly different AP performance.

We compare our method to the following baselines: Hotelling's $T^2$ method~\cite{macgregor1995statistical} and kernel density estimation
operating on single data points in the time series and learned with all of the points.
To allow for AP computation and a fair comparison, we use multiple thresholds on the scores to group single
point detections into intervals. All of the obtained intervals are then filtered with non-maximum-suppression (NMS) \cite{neubeck2006efficient} and receive as a score
the minimum value of the single-point detection score estimated by the baseline.

All of the methods evaluated use a time-delay preprocessing of $k=3$ and a subsequent NMS to obtain the 5 best scored non-overlapping intervals. Our MDI methods optimize over all possible intervals with sizes from $10$ to $50$. 

\vspace{-2.5mm}
\paragraph{Synthetic dataset}
We first test our algorithms on a synthetic dataset comprised of functions sampled from a Gaussian process prior (Gaussian kernel, $\sigma=1$).
The time series are perturbed at randomly sampled
intervals with sizes ranging from $5\%$ to $20\%$ of the length of the whole time series, which is set
to $250$. We simulate the following types of anomalies: (1) mean shift (MS): we set $x_t' = x_t - \mu$ with $\mu \in [3,4]$
within the anomaly region, (2) mean shift hard (MSH): MS with $\mu \in [0.5, 1]$, (3) amplitude change (AC): multiplying one dimension of the data points with
$1+g(t)$ with $g$ being a Gaussian window centered in the interval and with $2 \sigma$ matching the interval length, (4) frequency change (FC):
one dimension of the time series is sampled from a non-stationary Gaussian process prior~\cite{paciorek2004nonstationary} with
a change of the kernel hyperparameter within the anomaly interval. 
All of these types have $20$ univariate (MS, MSH, AC, FC) and $20$ multivariate ($D=5$) instances (MS$^5$, AC$^5$, FC$^5$) in the dataset.  
Note that the anomaly interval can only be detected in one dimension of the multivariate instances and our algorithms do not
have information about the dimension. 
We will release the code for generating the dataset and code for our algorithms to ensure reproducibility.

\begin{table}[tb]
    \caption{Results of our synthetic experiment for several baselines and several methods
    derived from our maximum divergent interval (MDI) framework. We use average precision (AP) and area under the ROC curve (AUC)
    as performance measures.}
    \label{tab:synthetic}
    \centering
    \resizebox{\linewidth}{!}{
        \input{tables/table_ap.tex}
    }
    \resizebox{\linewidth}{!}{
        \input{tables/table_auc.tex}
    }
\end{table}

\vspace{-3mm}
\paragraph{Results of our synthetic experiments}
The results of our synthetic experiments are given in Tab.~\ref{tab:synthetic} for AP and AUC performance. As can be seen from the AP results, the best
method is MDI Gaussian with a full model for the covariance matrices. MDI KDE is not able to deal with
frequency changes, since the correlations between the dimensions of subsequent data points are not taken
into account. This also holds for the ``no cov.'' and ``shared cov'' versions of MDI Gaussian. MDI Gaussian ``full cov.'' is also clearly the best method with respect to AUC
performance, however, there are a lot of cases where the AUC performance value of another method would not reveal the nearly random detection performance
measured by AP.

\vspace{-3mm}
\paragraph{Application of MDI to real datasets}
Meteocean  data (significant wave height, $H_{s}$, wind speed, $W$ and sea level pressure $SLP$) in a location near the Bahamas in the Atlantic Sea (23.838 N, 68.333 W) were used in these tests. Six months of hourly data, from June 2012 until November 2012 were extracted from the National Data Buoy Center from the NOAA\footnote{http://www.ndbc.noaa.gov/}. This period corresponds to the Atlantic hurricane season, which in that year was specially active with 19 tropical cyclones (winds above 52 km/h) were 10 of them became hurricanes (winds above 64 km/h).  
In contrast to our synthetic dataset, the anomalies have an effect on multiple variables at once.

\begin{figure}
  \centering
  \begin{minipage}[c]{0.5\textwidth}
     \begin{center}
    
    \includegraphics[width=.99\textwidth]{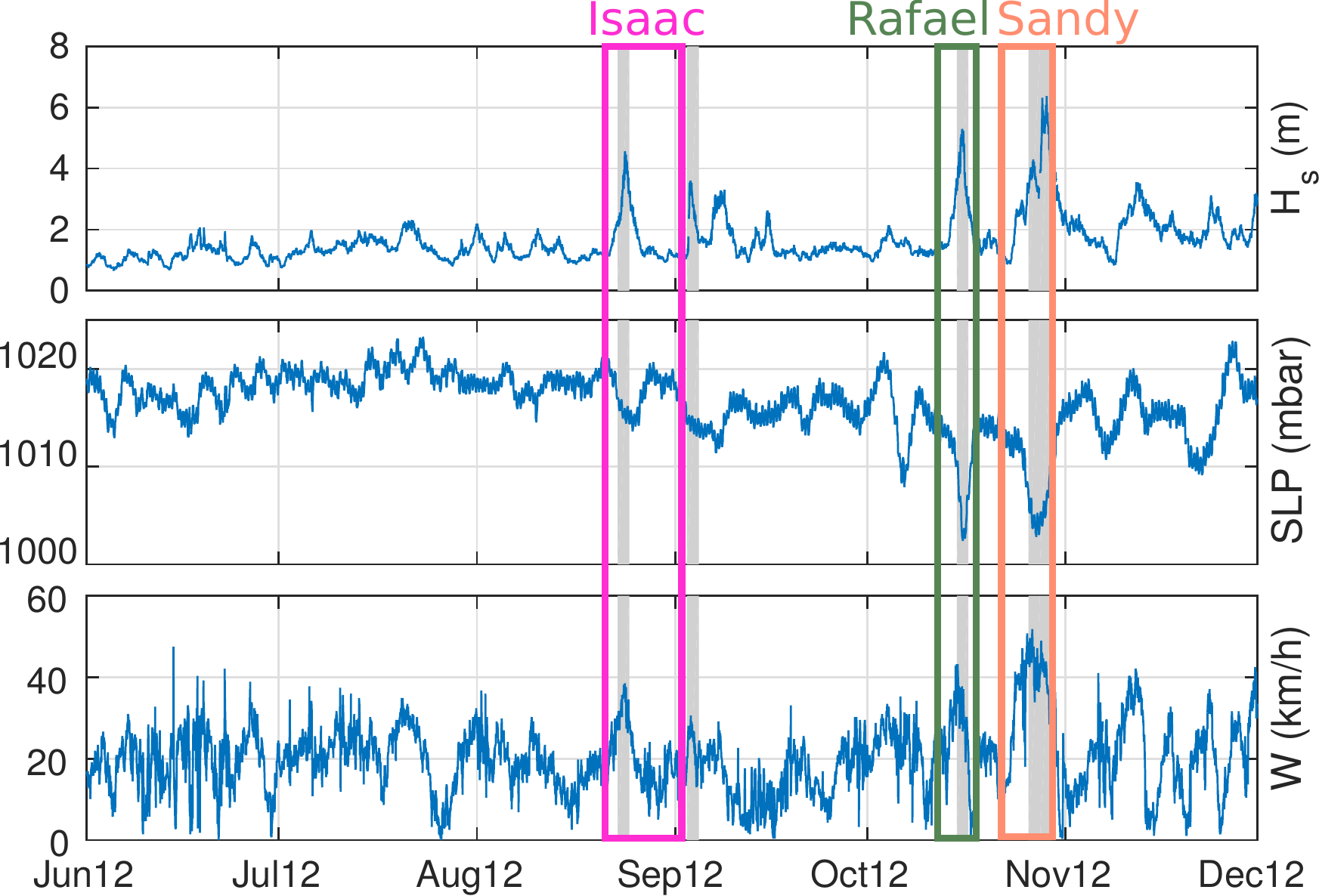}  
    \end{center}
  \end{minipage}
  \hfill
  \caption{ Boxes in colors represent historical hurricanes \emph{Isaac}, \emph{Rafael} and \emph{Sandy} and grey shaded areas MDI Gaussian detections. The false-positive right after Isaac might be related either to a local storm or to the reminiscences from hurricane \emph{Leslie} passing these days by Bermudas. } \label{fig:Res}
 
\end{figure}

We have applied the MDI Gaussian method to these three variables and compared the results with the historical hurricanes at Bahamas (Figure \ref{fig:Res}). The boxes in color represent the official duration of the three main events of that season that passed near our location, hurricanes \emph{Isaac}, \emph{Rafael} and \emph{Sandy} respectively. Grey shaded areas represent the MDI intervals detected by the model. Note that in general the ground-truth areas are larger than the detections, because they span the entire lifetime of the hurricane and not just its presence at the Bahamas.

\vspace{-3mm}
\section{Discussion and conclusions} \label{section:conclus}

We presented methods to detect anomalies in time series. All of our methods maximize a KL divergence
criterion that allows for finding intervals in time series that significantly differ from the rest with respect to their data distribution. 
We propose several variants for modeling the data distribution (kernel density estimation and different Gaussian assumptions) and analyze
their particular benefits and drawbacks in experiments.
In summary, our methods allow for efficient batch detection of anomalies in multivariate time series and are a useful tool
for data discovery in the natural sciences. 
Future work will be focused on automatically inferring the number of anomalous intervals.

\vspace{-3mm}
\paragraph{Acknowledgements}
The support of the project EU H2020-EO-2014 project BACI 'Detecting changes in essential ecosystem and biodiversity properties-towards a Biosphere Atmosphere Change Index, contract 640176 is gratefully acknowledged. 

\includegraphics[width=2cm]{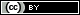} This work is licensed under a 
\href{http://creativecommons.org/licenses/by/3.0/}{Creative Commons Attribution 3.0 Unported License}.

\bibliography{MDRpaper}
\bibliographystyle{icml2016}

\end{document}

%% file: tables/table_ap.tex
\begin{tabular}{lccccccc}
\toprule
\textbf{Method/AP} & \textbf{MS} & \textbf{MSH} & \textbf{AC} & \textbf{FC} & \textbf{MS$^5$} & \textbf{FC$^5$} & \textbf{AC$^5$}\\
\midrule
Hotelling's $T^2$ test (pointwise) & 0.88 & 0.07 & 0.12 & 0.18 & 0.10 & 0.16 & 0.06\\
KDE (pointwise) & 0.90 & 0.10 & 0.13 & 0.00 & 0.18 & 0.04 & 0.29\\
Ours, MDI KDE & 0.97 & 0.12 & 0.20 & 0.00 & 0.82 & 0.00 & 0.43\\
Ours, MDI Gaussian (full cov.)& \textbf{1.00} & \textbf{0.44} & \textbf{0.79} & \textbf{1.00} & \textbf{1.00} & \textbf{0.82} & \textbf{0.62}\\
Ours, MDI Gaussian (no cov.) & 0.84 & 0.14 & 0.02 & 0.00 & 0.45 & 0.01 & 0.19\\
Ours, MDI Gaussian (shared cov.) & 0.32 & 0.06 & 0.01 & 0.01 & 0.10 & 0.04 & 0.04\\
\bottomrule
\end{tabular}

%% file: tables/table_auc.tex
\begin{tabular}{lccccccc}
\toprule
\textbf{Method/AUC} & \textbf{MS} & \textbf{MSH} & \textbf{AC} & \textbf{FC} & \textbf{MS$^5$} & \textbf{FC$^5$} & \textbf{AC$^5$}\\
\midrule
Hotelling's $T^2$ test (pointwise) & 0.98 & 0.58 & 0.92 & 0.94 & 0.72 & 0.82 & 0.80\\
KDE (pointwise) & 0.98 & 0.56 & 0.77 & 0.72 & 0.82 & 0.61 & 0.82\\
Ours, MDI KDE & 0.95 & 0.52 & 0.65 & 0.39 & 0.90 & 0.34 & 0.71\\
Ours, MDI Gaussian (full cov.)& \textbf{1.00} & \textbf{0.76} & \textbf{0.94} & \textbf{0.97} & \textbf{0.99} & \textbf{0.90} & \textbf{0.87}\\
Ours, MDI Gaussian (no cov.) & 0.90 & 0.54 & 0.69 & 0.46 & 0.87 & 0.45 & 0.75\\
Ours, MDI Gaussian (shared cov.) & 0.90 & 0.62 & 0.80 & 0.58 & 0.79 & 0.65 & 0.71\\
\bottomrule
\end{tabular}